\documentclass[10pt,twocolumn,letterpaper]{article}

\usepackage{iccv}
\usepackage{times}
\usepackage{epsfig}
\usepackage{graphicx}
\usepackage{amsmath}
\usepackage{amssymb}

\usepackage{booktabs}
\usepackage{multirow}
\usepackage{algorithm2e}
\usepackage[accsupp]{axessibility}

\usepackage[breaklinks=true,bookmarks=false]{hyperref}

\iccvfinalcopy % *** Uncomment this line for the final submission

\def\iccvPaperID{****} % *** Enter the ICCV Paper ID here

\ificcvfinal\fi

\begin{document}

%%%%%%%%% TITLE
\title{Learning Support and Trivial Prototypes for Interpretable Image Classification}

% \author{First Author\\
% Institution1\\
% Institution1 address\\
% {\tt\small firstauthor@i1.org}
% % For a paper whose authors are all at the same institution,
% % omit the following lines up until the closing ``}''.
% % Additional authors and addresses can be added with ``\and'',
% % just like the second author.
% % To save space, use either the email address or home page, not both
% \and
% Second Author\\
% Institution2\\
% First line of institution2 address\\
% {\tt\small secondauthor@i2.org}
% }

\author{
\parbox{1.0\linewidth}{\centering Chong Wang\textsuperscript{\rm 1} $\quad$ Yuyuan Liu\textsuperscript{\rm 1} $\quad$ Yuanhong Chen \textsuperscript{\rm 1} $\quad$ Fengbei Liu\textsuperscript{\rm 1} $\quad$ Yu Tian\textsuperscript{\rm 2} \\ $\quad$ $\quad$ $\quad$ $\quad$ Davis McCarthy\textsuperscript{\rm 3} $\quad$ Helen Frazer\textsuperscript{\rm 4} $\quad$ Gustavo Carneiro\textsuperscript{\rm 5} $\newline$   
\textsuperscript{\rm 1} Australian Institute for Machine Learning, University of Adelaide \\
\textsuperscript{\rm 2} Harvard University 
\textsuperscript{\rm 3} St Vincent's Institute of Medical Research \\
\textsuperscript{\rm 4} St Vincent's Hospital Melbourne 
\textsuperscript{\rm 5} CVSSP, University of Surrey
}}

\maketitle
% Remove page # from the first page of camera-ready.
\ificcvfinal\thispagestyle{empty}\fi

\begin{abstract}

Prototypical part network (ProtoPNet) methods have been designed to achieve interpretable classification by associating predictions with a set of training prototypes, which we refer to as trivial prototypes because they are trained to lie far from the classification boundary in the feature space. Note that it is possible to make an analogy between ProtoPNet and support vector machine (SVM) given that the classification from both methods relies on computing similarity with a set of training points (i.e., trivial prototypes in ProtoPNet, and support vectors in SVM). However, while trivial prototypes are located far from the classification boundary, support vectors are located close to this boundary, and we argue that this discrepancy with the well-established SVM theory can result in ProtoPNet models with inferior classification accuracy. In this paper, we aim to improve the classification of ProtoPNet with a new method to learn support prototypes that lie near the classification boundary in the feature space, as suggested by the SVM theory. In addition, we target the improvement of classification results with a new model, named ST-ProtoPNet, which exploits our support prototypes and the trivial prototypes to provide more effective classification. Experimental results on CUB-200-2011, Stanford Cars, and Stanford Dogs datasets demonstrate that ST-ProtoPNet achieves state-of-the-art classification accuracy and interpretability results. We also show that the proposed support prototypes tend to be better localised in the object of interest rather than in the background region. 

\end{abstract}

%%%%%%%%% BODY TEXT
% \vspace{-14pt}
\section{Introduction}
\vspace{-4pt}
\label{sec:intro}

Deep convolutional neural networks (CNN) \cite{krizhevsky2017imagenet,lecun2015deep,he2016deep} have had remarkable achievements in various visual tasks, e.g., image recognition \cite{he2016deep} and object detection \cite{ren2015faster}. Despite the excellent feature extraction and discrimination ability, CNNs are generally treated as black-box models due to their complex architectures, high-dimensional feature spaces, and the enormous number of learnable parameters. Such lack of interpretability hinders their successful application in fields that require understandable and transparent decisions \cite{rudin2019stop}, e.g., disease diagnosis \cite{tjoa2020survey,fang2019attention}, financial risk assessment \cite{liu2020predicting}, and autonomous driving \cite{kim2017interpretable}.

Recently, increasing attention has been dedicated to the development of interpretable deep-learning models \cite{koh2017understanding,alvarez2018towards,chen2019looks,bohle2022b}. 
A particularly interesting strategy is the prototype-based gray-box models, e.g., prototypical part network (ProtoPNet) \cite{chen2019looks,donnelly2022deformable}. 
These methods are inherently interpretable since they can explain the model's decisions by showing image classification activation maps associated with a set of class-specific image prototypes. 
These prototypes are automatically learned from training samples, with classification score being computed by comparing testing image parts to the learned training prototypes.

\begin{figure}[t!]
    \centering
    \includegraphics[width=1.00\linewidth]{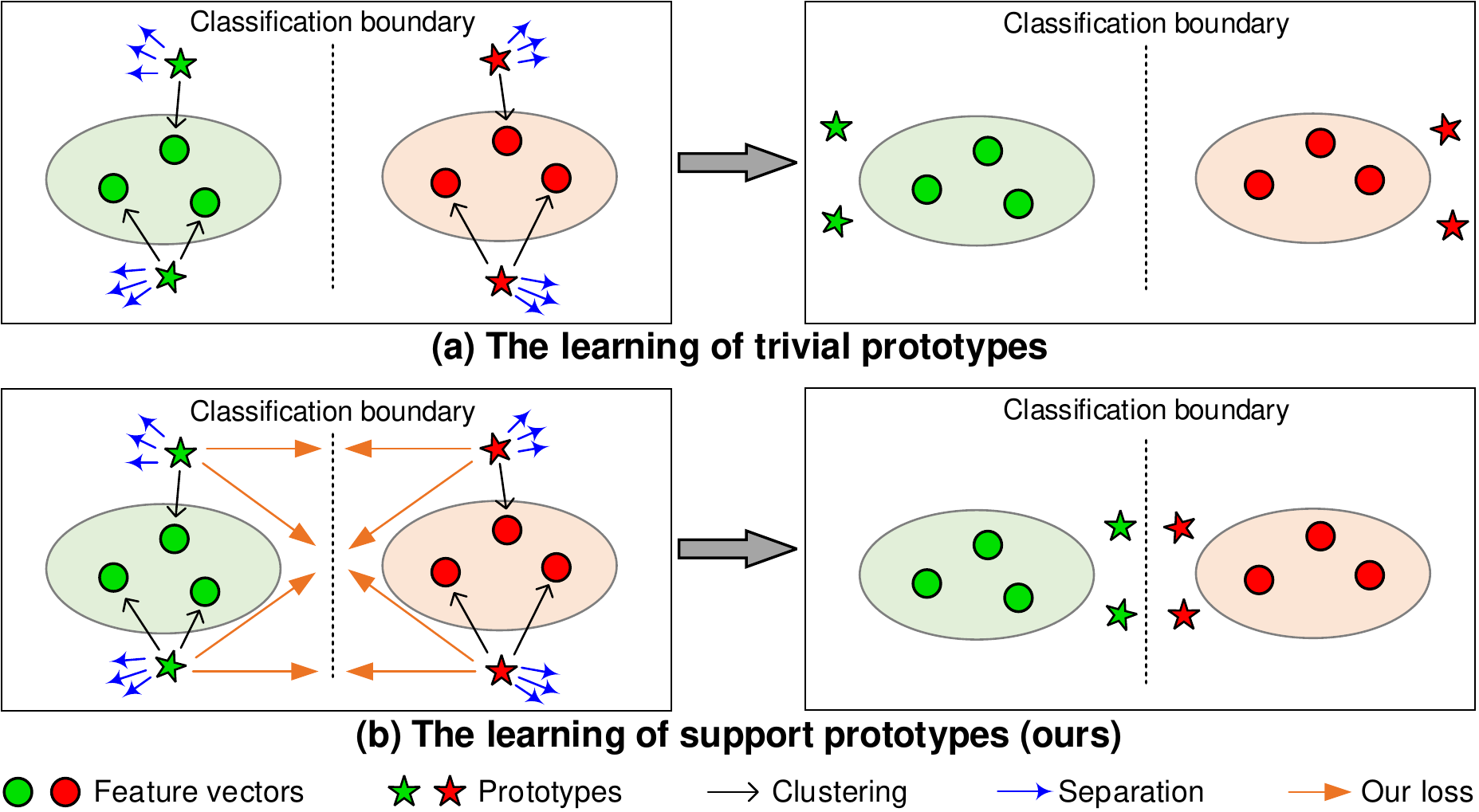}
    \vspace{-14pt}
    \caption{The difference between the learning of trivial and support prototypes. (a) Trivial prototypes: the separation loss pushes the prototypes of different classes as far as possible from the classification boundary. (b) Support prototypes: our new closeness loss enforces the prototypes of different classes to be as close as possible to the classification boundary.}
    \label{fig:protopnet}
    \vspace{-12.5pt}
\end{figure}

\begin{figure}[t!]
    \centering
    \includegraphics[width=1.00\linewidth]{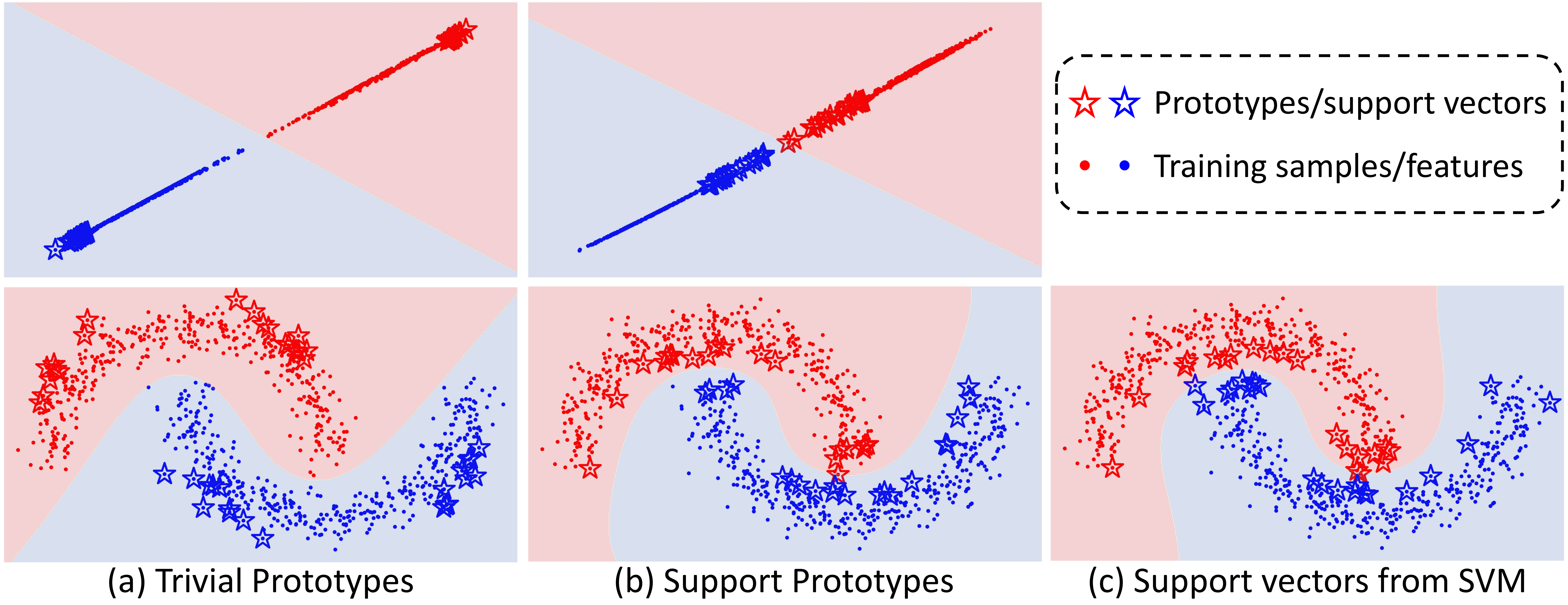}
    % \vspace{-15pt}
    \caption{Two-moon results. (a) Trivial prototypes and training samples in the feature (top) and data (bottom) spaces from the original ProtoPNet~\cite{chen2019looks}. (b) Support prototypes and training samples in the feature (top) and data (bottom) spaces from our method. (c) Support vectors and training samples from a Radial Basis Function (RBF) kernel based SVM~\cite{cortes1995support}. In (a) and (b), each prototype is projected onto the nearest training sample in the feature space. }
    \label{fig:twomoons}
    \vspace{-10pt}
\end{figure}

\begin{figure}[t!]
    \centering
    \includegraphics[width=1.00\linewidth]{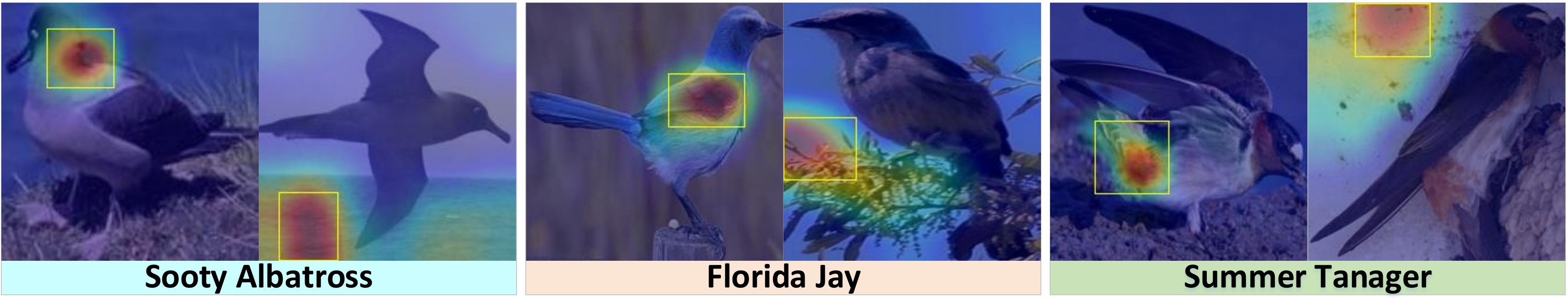}
    % \vspace{-20pt}
    \caption{Example prototypes sampled from a VGG19-based ProtoPNet~\cite{chen2019looks}. In each class, the left prototype focuses on object features while the right one captures background.}
    \label{fig:backgoundprototypes}
    \vspace{-10pt}
\end{figure}

ProtoPNet~\cite{chen2019looks} is trained to learn a classifier from a set of class-specific prototypes by minimising the cross-entropy classification loss and two additional regularisation losses, namely: 
1) a clustering loss that pulls together training image patches to at least one prototype of its own class; and 2) a separation loss that pushes apart training image patches from all prototypes of other classes.
%1) a clustering loss that pulls together training images to prototypes of their own class; and 2) a separation loss that pushes apart training images from all prototypes of other classes.
%More specifically, the clustering loss minimises the distance of each image patch to at least one prototype of its own class, while the separation loss maximises the distance between all image patches and all other class prototypes.
%the prototype (same-class) clustering and (different-class) separation losses are used to regularise the training of ProtoPNet, where the cluster loss encourages  each training image to have at least one local feature close to one prototype of its own class, and the separation loss enforces all local features of a training image to be far from the prototypes not of its own class, as shown in Fig.~\ref{fig:protopnet}. 
The combination of these two losses pushes the prototypes as far as possible from the classification boundary, but still within the class distribution, so we call them trivial prototypes, as shown in Fig.~\ref{fig:protopnet}(a).
We also display the trivial prototypes, learned with a feed-forward neural network\footnote{The network has an input layer of 2 nodes, a hidden layer of 256 nodes (activated by tanh), and an output layer of 2 nodes (activated by sigmoid).}, for the two-moon problem in Fig.~\ref{fig:twomoons}(a). 
Notice that these trivial prototypes are located far from the classification boundary.
In fine-grained visual classification, the trivial prototypes can mistakenly focus on %task-irrelevant 
background regions instead of on the object of interest~\cite{rymarczyk2021protopshare,rymarczyk2022interpretable}, particularly for those classes with subtle foreground (object) differences but large background variations, as shown in Fig.~\ref{fig:backgoundprototypes}. 
% \gustavo{The training of ProtoPNet and support vector machine (SVM)~\cite{cortes1995support} have strong similarities since both approaches depend on trivial prototypes and support vectors, respectively. Nevertheless, a noticeable difference is that while trivial prototypes lie far from the boundary, support vectors are located close to the classification boundary, as in Fig.~\ref{fig:twomoons}(c).}
Different from ProtoPNet's trivial prototypes, the support vector machine (SVM)~\cite{cortes1995support} classifier relies on a set of support vectors that are close to the classification boundary, as in Fig.~\ref{fig:twomoons}(c). These support vectors are often treated as hard samples.
% \chong{I think the first sentence has too strong link with SVM, so I prefer the second one.}
Motivated by SVM, we propose the derivation of support (i.e., hard-to-learn) prototypes for ProtoPNet methods.

In this paper, we propose an alternative learning strategy for ProtoPNet, which forces the learned prototypes to resemble SVM's support vectors and to be located as close as possible to the classification boundary.
% with the goal of increasing classification accuracy.
% to obtain more meaningful classification explanations. 
The strategy is formulated by a new closeness loss that minimises the distance between prototypes of different classes. 
As shown in Fig.~\ref{fig:protopnet}(b), our new loss enforces the prototypes to move closer to the classification boundary, as also demonstrated by Fig.~\ref{fig:twomoons}(b) revealing that the support prototypes produced by the introduction of our new closeness loss are indeed more similar to the support vectors of SVM in Fig.~\ref{fig:twomoons}(c). 
Furthermore, to improve the classification accuracy, we propose a new ST-ProtoPNet method that integrates both the support and trivial prototypes. 
The ST-ProtoPNet leverages the two distinct and complementary sets of prototypes to capture both hard (i.e., close to the boundary) and easy (i.e., far from the boundary) visual features for classification.
% Due to the different natures of the two sets of prototypes, they can also enable further improvements in terms of classification accuracy. 

%a ST-ProtoPNet framework that comprises an ensemble of two ProtoPNets, one using the classic formulation that produces the trivial prototypes, and another using our modified formulation which generates the support prototypes.
%To compensate the shortcomings of the trivial prototypes, in this paper we propose a ST-ProtoPNet which employs two sets of meaningful prototypes for interpretable image classification, namely trivial and support prototypes. The ST-ProtoPNet comprises a trivial ProtoPNet and a support ProtoPNet. 
%In this ST-ProtoPNet, while the trivial prototypes capture easy-to-learn visual patterns from training samples, the support prototypes are the hard-to-learn ones near the classification boundary in the feature space, as shown in Fig.~\ref{fig:twomoons}(b). 

The major contributions of this work are: 
\begin{enumerate}
    \vspace{-3pt}
    \item We provide the first study that makes an analogy between the prototype learning from ProtoPNet and support vector learning from SVM, where we propose support (i.e., hard-to-learn) prototypes that can improve classification accuracy and interpretability.
    \vspace{-3pt}
    \item We present a new ST-ProtoPNet method to exploit both support and trivial prototypes for interpretable image classification, where the two sets of prototypes can provide complementary information to improve classification accuracy. 
    \vspace{-3pt}
    \item We conduct extensive experiments on three benchmarks, showing that our ST-ProtoPNet outperforms current state-of-the-art (SOTA) methods in terms of classification accuracy and interpretability. 
    \vspace{-3pt}
\end{enumerate}
In our experiments, we also demonstrate that the trivial and support prototypes have different characteristics,
where the trivial prototypes tend to focus on both local parts of the visual object of interest and the background, while the support prototypes mainly focus on object parts belonging to the visual class of interest. 

% \begin{figure}[t!]
%     \centering
%     \includegraphics[width=1.00\linewidth]{figures/Fig.2.pdf}
%     % \vspace{-15pt}
%     \caption{Two-moon results. (a) Trivial prototypes and training samples in the feature (top) and data (bottom) spaces from the original ProtoPNet~\cite{chen2019looks}. (b) Support prototypes and training samples in the feature (top) and data (bottom) spaces from our method. (c) Support vectors and training samples from a Radial Basis Function (RBF) kernel based SVM~\cite{cortes1995support}. In (a) and (b), each prototype is projected onto the nearest training sample in the feature space. \chong{I have changed the middle figure to make it look differently from the trivial one. }}
%     \label{fig:twomoons}
%     \vspace{-10pt}
% \end{figure}

\section{Related Work}
\label{sec:relatedwork}

% In this section, we first review relevant studies on classification interpretability where we focus on prototype-based methods, and then we briefly review support vector machine (SVM) classification.
% Finally, we provide a short survey on ensemble classification for interpretability. 

\subsection{Classification Interpretability}

The interpretation of classification results produced by deep neural networks can be achieved by a variety of post-hoc explanation techniques, e.g., explanatory surrogates \cite{lundberg2017unified,zhang2018interpreting,shitole2021one}, counterfactual examples \cite{goyal2019counterfactual,teney2020learning,kenny2021generating}, and saliency visualisation \cite{simonyan2013deep,zeiler2014visualizing,zhou2016learning,selvaraju2017grad}. 
Alternatively, prototype-based interpretable techniques can access the model's inner computations.
%In comparison with post-hoc explanations, prototype-based interpretability is directly present in the model's inner computations. 
ProtoPNet~\cite{chen2019looks} is the original work that uses class-specific prototypes for interpretable image classification. 
Similar to ProtoPNet, TesNet \cite{wang2021interpretable} constructs class-specific transparent basis concepts on Grassmann manifold for the interpretable classification. 
Derived from ProtoPNet, Deformable ProtoPNet~\cite{donnelly2022deformable} employs spatially-flexible and deformable prototypes to adaptively capture meaningful object features. 
In ProtoPShare \cite{rymarczyk2021protopshare}, a data-dependent merge-pruning method is presented to share prototypes among classes, which can reduce the number of prototypes used for classification. 
In contrast, ProtoPool \cite{rymarczyk2022interpretable} introduces a fully differentiable prototype assignment strategy to reduce the number of prototypes.
In Proto2Proto \cite{keswani2022proto2proto}, a knowledge distillation method is designed to transfer interpretability from a teacher ProtoPNet to a shallow student ProtoPNet. 
ProtoTree \cite{nauta2021neural} integrates the prototype learning into a binary neural decision tree that can explain its predictions by tracing a decision path throughout the tree. 
ViT-NeT \cite{kim2022vit} further establishes the prototype neural tree structure on visual transformers \cite{dosovitskiy2020image}. 

Because of the ability to self-explain classification results, prototype-based interpretability (e.g., ProtoPNet) has been widely utilised not only in the computer vision applications above, but also in medical imaging \cite{barnett2021case,kim2021xprotonet,wang2022knowledge} and face recognition \cite{trinh2021interpretable}. 
However, an open question faced by these methods is if the prototypes being learned are the ideal ones in terms of classification and interpretability.

\vspace{-1pt}
\subsection{SVM vs Prototype-based Classification}
\vspace{-1pt}

To better understand the role 
% \chong{optimality}
of prototypes, we consider the support vector machine (SVM)~\cite{cortes1995support} classifier that finds support vectors to represent classes.
More specifically, SVM learns the maximum-margin classifier defined by a classification boundary that maximises the distance to the closest training samples, which are the support vectors for the classes.
The testing of SVM consists of computing a weighted similarity between a testing sample and the support vectors.
It is interesting to note that the testing of prototype-based classifiers is also based on measuring the similarity between a testing image and a set of class-specific prototypes learned from the training process. 
Although the testing of SVM and prototype-based classifiers are similar, their training procedures are different. 
First, the training of a prototype-based classifier learns a fixed number of prototypes~\cite{chen2019looks,donnelly2022deformable}, while the SVM classifier learns to weight a variable number of support vectors from the training set.
Second, in prototype-based classifiers, the learned prototypes tend to be far from the classification boundary, which is contrary to the SVM training objective mentioned above.

The study of deep learning methods from an SVM theoretical perspective is a rich area of research~\cite{domingos2020every,pruthi2020estimating,chen2021equivalence}, but 
there are many practical questions that need to be addressed, e.g., how to scale the kernel computation for large-scale datasets, how to shorten the training process~\cite{pruthi2020estimating}, and how to integrate  deep-learning features with the learning of the SVM classifier. 
In this paper, our focus is on adapting the learning of ProtoPNet's prototypes to make them similar to SVM's support vectors, by forcing prototypes to be as close as possible to the classification boundary.

\vspace{-1pt}
\subsection{Interpretable Ensemble Classification}
\vspace{-1pt}

Ensemble classification~\cite{dong2020survey} is a classical machine learning approach that combines the results from multiple classifiers, with the goals of improving learning generalisation and classification calibration. 
The use of interpretable ensemble strategy 
%to improve the classification accuracy 
has been explored in \cite{chen2019looks,wang2021interpretable,nauta2021neural,donnelly2022deformable,rymarczyk2022interpretable}, which is achieved by summing the classification logits of multiple prototype-based classifiers (e.g., ProtoPNets trained with different CNN backbones). 
In this work, we propose an interpretable ensemble classification by combining the predictions of two ProtoPNets with highly distinctive prototypes (i.e., support and trivial prototypes), which is different from previous studies where the type of prototypes produced by each classifier is very similar given that the same training objective is used for each classifier. 
% More specifically, the ensemble classification used in this paper targets the utilisation of two complementary sets of prototypes, represented by the support and trivial prototypes.
%, particularly when the prototypes are learned from quite different objective functions, such as the ones for learning the support and trivial prototypes. 

%\gustavo{We can then say that even though they have multiple backbones, the type of prototypes that each model produces are very similar given that the objective function is the same. That is, they may improve classification results, but not interpretability results. We need to find a way to measure interpretability -- maybe with the same measure you used in the MICCAI paper that measures the diversity of the prototypes?}\chong{That is a good idea. I have re-organised this section.}

% \gustavo{I think this section is too short.  I just want to say that we're the first to propose ensemble for interpretability.  I saw one paper online about this, but they basically train many classifiers and show multiple CAM maps, so I'm not counting that., here is the paper: Jiang, Hongyang, Kang Yang, Mengdi Gao, Dongdong Zhang, He Ma, and Wei Qian. "An interpretable ensemble deep learning model for diabetic retinopathy disease classification." In 2019 41st annual international conference of the IEEE engineering in medicine and biology society (EMBC), pp. 2045-2048. IEEE, 2019.}
% \chong{how about we ignore this paper? because it doesn't use prototypes.}

\vspace{-3pt}
\section{Preliminaries}
\label{sec:preliminaries}
\vspace{-3pt}

We assume to have a training set $\mathcal{D} = \{ (\mathbf{x}_n,\mathbf{y}_n) \}_{n=1}^{|\mathcal{D}|}$,
where $\mathbf{x} \in \mathcal{X} \subset \mathbb{R}^{H \times W \times R}$ is an image with $R$ colour channels and $\mathbf{y} \in \mathcal{Y} \subset \{0,1\}^C$ is a one-hot vector representation of the image class label.
The interpretable ProtoPNet \cite{chen2019looks,donnelly2022deformable} is trained to learn a set of prototypes $\mathcal{P}=\{\mathbf{p}_m\}_{m=1}^{M}
$, where $\mathbf{p}_m \in \mathbb{R}^ {\rho_1 \times \rho_2 \times D}$,  
with each of the $C$ classes containing $M/C$ prototypes.
Without loss of generality, we assume  $\rho_1=\rho_2=1$, but the extension to general values is trivial.
%\gustavo{are $\rho_1$, $\rho_2$ and $d$ fixed or variable for the $M$ prototypes?\chong{fixed, they are hyperparameters}}
A typical ProtoPNet comprises four components: a CNN backbone, add-on layers, a prototype layer, and a fully connected (FC) layer. 
An input image $\mathbf{x}$ is fed to the CNN backbone $f_{\theta}:\mathcal{X} \to \mathcal{F}$ (parameterised by $\theta \in \Theta$, where $\mathcal{F} \subset \mathbb{R}^{ \bar{H} \times \bar{W} \times \bar{D}}$) and then passed on to the add-on layers, denoted by $f_{\omega}:\mathcal{F} \to \mathcal{V}$ (parameterised by $\omega \in \Omega$), to produce a feature map $\mathbf{V} \in \mathcal{V} \subset \mathbb{R}^{\bar{H} \times \bar{W}  \times D}$. 
% (with $\bar{H} < H$, $\bar{W} < W$, and $\bar{D} < D$),
%\gustavo{$h < H$, $w < W$?}\chong{yes, to produce feature map $\mathbf{V} \in \mathcal{V} \subset \mathbb{R}^{h \times w \times d}$ ($h < H$, $w < W$)}, 
% whose channel number is matched with the prototype dimension $D$ \gustavo{how are they matched? maybe we can ignore this matching process?}.
% \chong{In ProtoPNet, the prototype length $d$ is a pre-defined hyperparameter, typically 64 in our work. For a CNN, we can have high-dimensional (512 in Res34, 2048 in DenseNet121) feature maps from the last conv layer. Since we need to compute the similarity between the prototypes and feature maps, they must have the same dimension, the add-on layers on the top of CNN backbone can do this by reducing the channel of feature maps and make it match with the prototype dimension $d$. }
% \gustavo{Can you add a short explanation here?}
The prototype layer computes the similarity between the feature map $\mathbf{V}$ and the $M$ $D$-dimensional prototypes $\{ \mathbf{p}_m\}_{m=1}^{M} $ to generate $M$ similarity maps $\mathbf{S}_m^{(i,j)} = {\rm{sim}}(\mathbf{V}(i,j,:), \mathbf{p}_m)$, where 
$i \in \{1,...,\bar{H}\}$, 
$j \in \{1,...,\bar{W}\}$,
and $\rm{sim(\cdot,\cdot)}$ represents a similarity measure, e.g., cosine similarity \cite{donnelly2022deformable} and projection metric \cite{wang2021interpretable}. 
The prototype layer outputs $M$ similarity scores from max-pooling $\mathcal{S} = \Big\{\underset{i \in \{1,...,\bar{H}\}, j \in \{1,...,\bar{W}\}}\max \; \mathbf{S}_m^{(i,j)}\Big\}_{m=1}^M$, which are fed to the FC layer $f_{\phi}:\mathcal{S} \to \Delta$, parameterised by $\phi \in \Phi$, 
to produce the classification prediction $\hat{\mathbf{y}} \in \Delta \subset [0,1]^C$, where $\Delta$ denotes the probability space for $C$ classes.

%\chong{In our work, $\rho_1=\rho_2=1$, which means our prototypes are vectors. I introduce the two variables in Preliminaries to make it general, because the Deformable ProtoPNet uses 2 * 2 prototypes. Can we say we use 1 * 1 here, and omit the two variables in the following parts for simplicity? }\gustavo{We can say, without loss of generality, we assume that $\rho_1=\rho_2=1$, where the extension to larger values for $\rho_1$ and $\rho_2$.  Can you update this part?} 

\vspace{-3pt}
\section{ST-ProtoPNet}
\vspace{-3pt}

\begin{figure}[t!]
    \centering
    \includegraphics[width=1.00\linewidth]{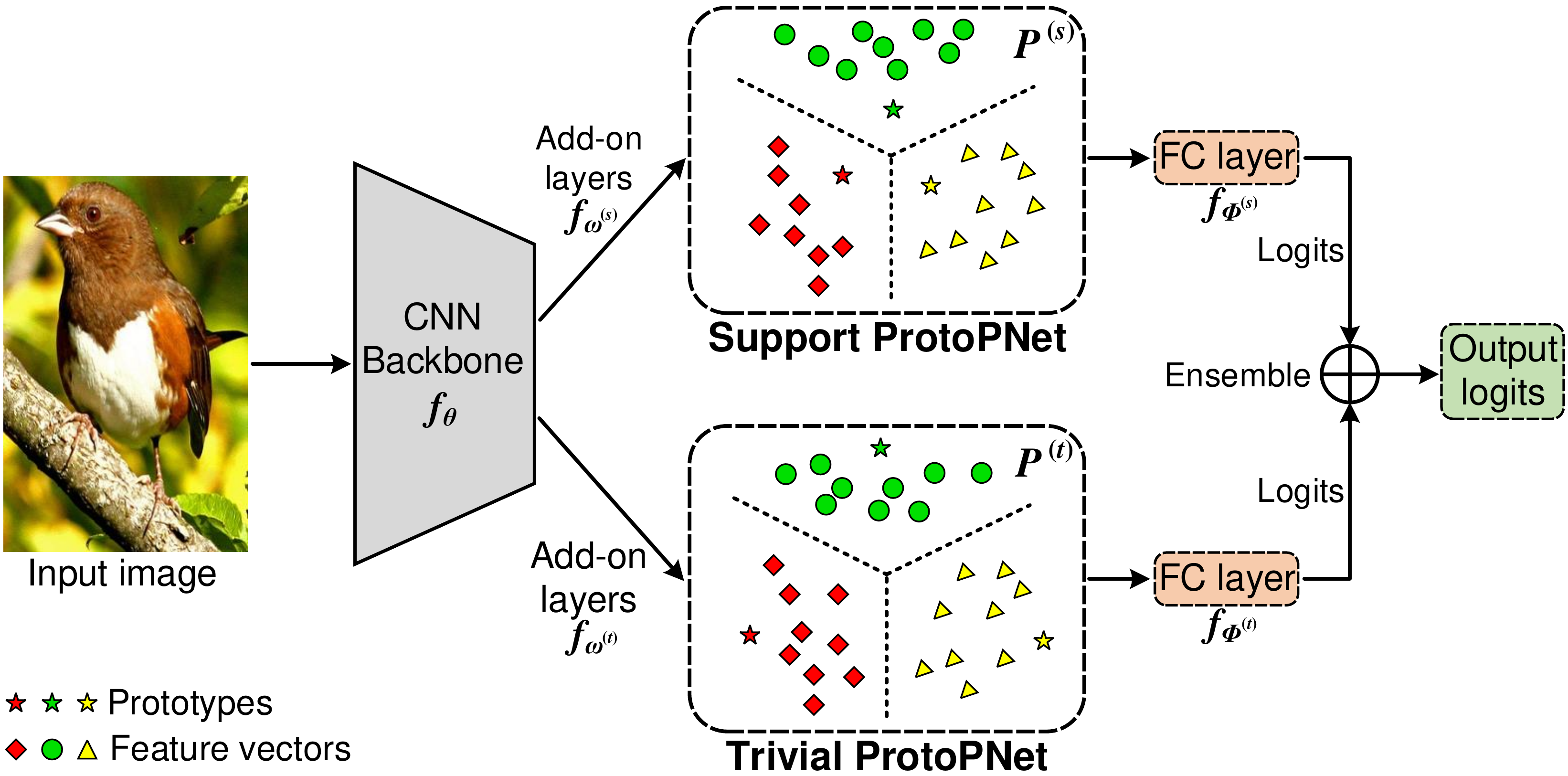}
    % \vspace{-20pt}
    \caption{The architecture of our proposed ST-ProtoPNet method for the interpretable image classification.}
    \label{fig:archetecture}
    \vspace{-10pt}
\end{figure}

An overview of our proposed ST-ProtoPNet method is illustrated in Fig.~\ref{fig:archetecture}, which comprises a shared CNN backbone $f_{\theta}(\cdot)$, two interpretable ProtoPNet classification branches, namely: 1) the support ProtoPNet
represented by add-on layers $f_{\omega^{(s)}}(\cdot)$, prototype layer with support prototypes $\mathcal{P}^{(s)}$, and FC layer $f_{\phi^{(s)}}(\cdot)$ which outputs the classification probability distribution $\hat{\mathbf{y}}^{(s)} \in \Delta$; and 2) the trivial ProtoPNet branch with its add-on layers $f_{\omega^{(t)}}(\cdot)$, trivial prototypes $\mathcal{P}^{(t)}$, and FC layer $f_{\phi^{(t)}}(\cdot)$ that generates probability predictions $\hat{\mathbf{y}}^{(t)} \in \Delta$.
The final classification is obtained by combining the classification logits from both the support and trivial ProtoPNets. 
In our implementation, we construct the support and trivial ProtoPNet mainly based on the original ProtoPNet \cite{chen2019looks} and TesNet \cite{wang2021interpretable}, as explained below.

\subsection{Support ProtoPNet}
\label{sec:support}

The support ProtoPNet is designed to produce support %(i.e., hard-to-learn) 
prototypes %for classification 
that are as close as possible to the classification boundary, as shown in Fig.~\ref{fig:protopnet}(b) and~\ref{fig:twomoons}(b).
%, which is the opposite goal from the trivial ProtoPNet that produces prototypes that are as far as possible from the classification boundary, as shown in Fig.~\ref{fig:protopnet} and Fig.~\ref{fig:twomoons}.
%The trivial ProtoPNet only employs a set of easy-to-learn prototypes for classification, which cannot fully capture the complicated visual patterns of training samples. To compensate the shortcomings of the trivial ProtoPNet, we further propose a support ProtoPNet to learn hard training prototypes, which can provide complementary information for the trivial ProtoPNet. 
%Our support ProtoPNet is inspired by support vector machines (SVMs) \cite{cortes1995support} in which the hard-to-learn samples are exploited to support the classification boundary. 
%In order to obtain support prototypes, we explicitly enforce the prototypes of different classes to be close to each other. The overall training objective of the support ProtoPNet is defined as follows:
The loss function to optimise the support ProtoPNet branch is defined as: 
\begin{equation}
\begin{split}
    \theta^*,&\omega^{(s)*},\mathcal{P}^{(s)*},\phi^{(s)*} = \\ &\arg\min_{\theta,\omega^{(s)},\mathcal{P}^{(s)},\phi^{(s)}} \sum_{(\mathbf{x},\mathbf{y}) \in \mathcal{D}}\ell_{spt}(\mathbf{x},\mathbf{y},\theta,\omega^{(s)},\mathcal{P}^{(s)},\phi^{(s)}).
\end{split}
    \label{eq:main_loss_support}
\end{equation}
The loss for each training sample $(\mathbf{x},\mathbf{y}) \in \mathcal{D}$ in Eq.~\eqref{eq:main_loss_support} above is represented by: 
\begin{equation}
\begin{split}
    \ell_{spt}(\mathbf{x},\mathbf{y},\theta,\omega^{(s)},\mathcal{P}^{(s)},\phi^{(s)}) = \ & \ \ell_{ce}(\mathbf{x},\mathbf{y},\theta,\omega^{(s)},\mathcal{P}^{(s)},\phi^{(s)})  \\
    &- \lambda_1 \ell_{ct}(\mathbf{x},\mathbf{y}, \theta,\omega^{(s)},\mathcal{P}^{(s)})  \\
    &+ \lambda_2 \ell_{sp}(\mathbf{x},\mathbf{y}, \theta,\omega^{(s)},\mathcal{P}^{(s)})  \\
    &- \lambda_3 \ell_{cls}(\mathcal{P}^{(s)})  \\
    &+ \lambda_4 \ell_{ort}(\mathcal{P}^{(s)}),
\end{split}
\label{eq:support_PPN}
\end{equation}
where $\lambda_1$, $\lambda_2$, $\lambda_3$, and $\lambda_4$ are hyper-parameters to balance each term, $\ell_{ce}(\cdot)$ denotes the cross-entropy classification loss, $\ell_{ct}(\cdot)$ and $\ell_{sp}(\cdot)$ represent the clustering and separation losses, respectively, which are introduced to regularise the ProtoPNet's training, as follows:
\begin{equation}
    \ell_{ct}(\mathbf{x},\mathbf{y}, \theta,\omega^{(s)},\mathcal{P}^{(s)}) = \max_{\mathbf{p}\in\mathcal{P}^{(s)}_{\mathbf{y}}} \max_{\mathbf{v} \in\mathbf{V}^{(s)}}  \rm{sim}(\mathbf{v},\mathbf{p}),
    \label{eq:support_ct}
\end{equation}
\begin{equation}
    \ell_{sp}(\mathbf{x},\mathbf{y}, \theta,\omega^{(s)},\mathcal{P}^{(s)}) = \max_{\mathbf{p}\notin\mathcal{P}^{(s)}_{\mathbf{y}}} \max_{\mathbf{v} \in\mathbf{V}^{(s)}}  \rm{sim}(\mathbf{v},\mathbf{p}),
    \label{eq:support_sp}
\end{equation}
where $\mathbf{V}^{(s)} = f_{\omega^{(s)}}(f_{\theta}(\mathbf{x}))$ is the feature map extracted from the input image $\mathbf{x}$, 
$\mathbf{v}$ represents one of the $\bar{H} \times \bar{W}$ feature vectors in $\mathbf{V}^{(s)}$ obtained by matrix vectorisation, 
$\mathbf{p}$ is a normalised prototype (i.e., unit vector) in $\mathcal{P}^{(s)}$, 
${\rm{sim}}(\cdot,\cdot)$ is one of the similarity functions defined in Sec.~\ref{sec:preliminaries},
and $\mathcal{P}^{(s)}_{\mathbf{y}}$ denotes the set of prototypes of class $\mathbf{y}$.
The clustering loss in Eq. \eqref{eq:support_ct} and separation loss in Eq. \eqref{eq:support_sp} aim to learn a meaningful feature space in which the image features of a certain class are clustered around the prototypes of the class, and also well separated from those of other classes. 

As mentioned in Sec.~\ref{sec:intro}, the effect of the clustering and separation losses above tend to push the prototypes of different classes as far as possible from the classification boundary, resulting in trivial prototypes, as displayed in Fig.~\ref{fig:protopnet}(a) and Fig.~\ref{fig:twomoons}(a). In order to learn the proposed support prototypes, we introduce the following novel closeness loss $\ell_{cls}$ to explicitly enforce the prototypes of different classes to be close to each other, which is formulated as:
\begin{equation}
    % \ell_{cls}(\mathcal{P}^{(s)}) = \sum_{c_1 = 1}^{C-1} \sum_{c_2 = c_1 + 1}^{C} \min_{\mathbf{p}_m\in\mathcal{P}_{c_1}, \mathbf{p}_n\in\mathcal{P}_{c_2}} \frac{\mathbf{p}_m^{\top} \mathbf{p}_n}{\|\mathbf{p}_m\|\|\mathbf{p}_n\|},
    % \ell_{cls}(\mathcal{P}^{(s)}) = \sum_{c_1 = 1}^{C-1} \sum_{c_2 = c_1 + 1}^{C} \min_{\mathbf{p}_m\in\mathcal{P}_{c_1}, \mathbf{p}_n\in\mathcal{P}_{c_2}} \rm{sim}(\mathbf{p}_m, \mathbf{p}_n).
    \ell_{cls}(\mathcal{P}^{(s)}) = \sum_{c_1 = 1}^{C-1} \sum_{c_2 = c_1 + 1}^{C} \min_{\mathbf{p}_m\in\mathcal{P}_{c_1}, \mathbf{p}_n\in\mathcal{P}_{c_2}} 
    \mathbf{p}_m^{\top} \mathbf{p}_n.
    \label{eq:support_cls}
\end{equation} 
%\chong{I think it is better to directly use dot product in Eq.~\eqref{eq:support_cls}, because it is straightforward. }

During training, this closeness loss $\ell_{cls}$ maximises the pair-wise prototype similarity, in the form of dot product $\mathbf{p}_m^{\top} \mathbf{p}_n$ between different classes in Eq.~\eqref{eq:support_cls} above, with the goal of pulling the prototypes close to the classification boundary. 
On the one hand, as the prototypes move gradually towards the classification boundary, they are able to capture harder visual features from training samples. 
On the other hand, since the prototypes are located near the classification boundary, they can put pressure on the support ProtoPNet's feature learning (i.e., enforce it to learn highly discriminative 
feature representations for accurate classification), which is beneficial to extract more meaningful semantic information from training samples.

Ideally, each prototype of a class should focus on unique object parts of the training images (e.g., head, tail, and claw of birds), so that the prototypes can represent rich and diverse visual patterns. 
However, there is no particular constraints to guarantee such prototype diversity and the issue of prototype duplication \cite{donnelly2022deformable} often occurs in the ProtoPNet family of models. 
To encourage the intra-class prototype diversity, we employ an orthonormality loss \cite{wang2021interpretable} so that prototypes within a class can represent dissimilar visual patterns of training samples, which is defined as: 
\begin{equation}
    \ell_{ort}(\mathcal{P}^{(s)}) = \sum_{c = 1}^{C} \| {\mathbf{P}_c}^{\top} \mathbf{P}_c -\mathbf{I}_{M/C} \|_F^2,
    \label{eq:support_orth}
\end{equation}
where $\|\cdot\|_F^2$ represents Frobenius norm, 
$\mathbf{P}_c \in D \times \mathbb{R}^{ (M/C)}$ stands for a matrix composed of the prototypes of class $c$ (prototypes in each column of $\mathbf{P}_c$ are normalised), 
and $\mathbf{I}_{M/C}$ is an identity matrix of size $M/C \times M/C$. 

% \gustavo{even though we mentioned many types of similarity functions, these loss functions rely on dot product with a normalised $\mathbf{v}$. Should we say $\rm{sim(\cdot,\cdot)}$ here or leave this dot product? Is $\mathbf{p}$ normalised?}
% \chong{yes, $\mathbf{p}$ is normalised in all our experiments}.
% \chong{For the similarity function, we actually use different forms for cropped and full datasets. In any form, the prototypes are always normalised. 
% Cropped (Table 1): following TesNet, dot product between \textbf{un-normalised} feature map and prototypes as similarity score for classification, while dot product between normalised feature map and prototypes in Eq. \eqref{eq:support_ct} and Eq. \eqref{eq:support_sp}. 
% Full (Table 2): following Deformable ProtoPNet, dot product between normalised feature map and prototypes as similarity score for classification, and dot product between normalised feature map and prototypes in Eq. \eqref{eq:support_ct} and Eq. \eqref{eq:support_sp}. }
% \chong{Can we give these details in the experimental section, and just say we use dot product and modify Eq. \eqref{eq:support_ct} and Eq. \eqref{eq:support_sp} in the method section? }
% \gustavo{In this case, I prefer to say sim(.,.) here and give the details in the experiments.}

\subsection{Trivial ProtoPNet}

%\chong{ The word "stratified" is fine if we consider the hard and easy prototypes?}

As described in Sec.~\ref{sec:support}, the support ProtoPNet is developed to learn support (i.e., hard-to-learn) prototypes by forcing them to be close to the classification boundary. 
Considering that training samples contain not only hard visual features but also important easy ones that the support prototypes cannot completely capture, we propose to also learn trivial prototypes to provide complementary classification information, and exploit both the support and trivial prototypes for improved interpretable classification.
% that relies on two highly distinctive sets of prototypes. 
%  relies on two much richer sets of visually distinctive prototypes.
% \gustavo{I added a short explanation "improved interpretable classification that relies on a much richer set of visually diverse prototypes".  I wouldn't know what else to add...}

The loss objective to optimise the trivial ProtoPNet branch is defined as follows: 
\begin{equation}
\begin{split}
    \theta^*,&\omega^{(t)*},\mathcal{P}^{(t)*},\phi^{(t)*} = \\ &\arg\min_{\theta,\omega^{(t)},\mathcal{P}^{(t)},\phi^{(t)}} \sum_{(\mathbf{x},\mathbf{y}) \in \mathcal{D}}\ell_{trv}(\mathbf{x},\mathbf{y},\theta,\omega^{(t)},\mathcal{P}^{(t)},\phi^{(t)}).
\end{split}
    \label{eq:main_loss_trivial}
\end{equation}
The loss for each training image $(\mathbf{x},\mathbf{y}) \in \mathcal{D}$ in Eq.~\eqref{eq:main_loss_trivial} above is represented by: 
\begin{equation}
\begin{split}
    \ell_{trv}(\mathbf{x},\mathbf{y},\theta,\omega^{(t)},\mathcal{P}^{(t)},\phi^{(t)}) = \ & \ \ell_{ce}(\mathbf{x},\mathbf{y},\theta,\omega^{(t)},\mathcal{P}^{(t)},\phi^{(t)})  \\
    &- \lambda_1 \ell_{ct}(\mathbf{x},\mathbf{y}, \theta,\omega^{(t)},\mathcal{P}^{(t)})  \\
    &+ \lambda_2 \ell_{sp}(\mathbf{x},\mathbf{y}, \theta,\omega^{(t)},\mathcal{P}^{(t)})  \\
    &+ \lambda_3 \ell_{dsc}(\mathcal{P}^{(t)}) \\
    &+ \lambda_4 \ell_{ort}(\mathcal{P}^{(t)}),  
\end{split}
\label{eq:trivial_PPN}
\end{equation}
where $\lambda_1$, $\lambda_2$, $\lambda_3$, and $\lambda_4$ are hyper-parameters, 
$\ell_{ce}(\cdot)$ is the cross-entropy loss,
the clustering loss $\ell_{ct}$, separation loss $\ell_{sp}$, and orthonormality loss $\ell_{ort}$ are the same as in the support ProtoPNet defined in Eq. ~\eqref{eq:support_ct}, \eqref{eq:support_sp} and \eqref{eq:support_orth}, respectively.

The trivial ProtoPNet targets the learning of easy prototypes that are far from the classification boundary and have a good discrimination ability. 
To help achieve this, we introduce a new discrimination loss $\ell_{dsc}$ to facilitate the inter-class separability between prototypes of different classes. 
% \gustavo{is this a new loss?  Isn't that the loss used in common ProtoPNet methods?}. 
% \chong{This is a new loss proposed by us and not used in other ProtoPNet methods. In TesNet method, there is a similar loss also for prototype separability, but it has different form from ours.} 
% \gustavo{So, why do we use this new loss instead of the losses proposed before? We need to justify that.}
% \chong{This loss is different from \cite{wang2021interpretable} that uses intra-class outer product to obtain a matrix for quantifying prototype distances between classes. Can we say our proposed discrimination loss has a very similar form with the closeness loss?}
This is formulated by minimising the pair-wise prototype similarities of different classes, as follows:
\begin{equation}
    % \ell_{discri}(\mathcal{P}^{(s)})  = \sum_{c_1 = 1}^{C-1} \sum_{c_2 = c_1 + 1}^{C} \max_{\mathbf{p}_m\in\mathcal{P}_{c_1}, \mathbf{p}_n\in\mathcal{P}_{c_2}} \frac{\mathbf{p}_m^{\top} \mathbf{p}_n}{{\|\mathbf{p}_m\|\|\mathbf{p}_n\|}}.
    \ell_{dsc}(\mathcal{P}^{(t)})  = \sum_{c_1 = 1}^{C-1} \sum_{c_2 = c_1 + 1}^{C} \max_{\mathbf{p}_m\in\mathcal{P}_{c_1}, \mathbf{p}_n\in\mathcal{P}_{c_2}} \mathbf{p}_m^{\top} \mathbf{p}_n.
    \label{eq:ell_dis}
\end{equation}

% We may need to say a little things here.

\subsection{Training and Testing}

\textbf{Training.} Following the training strategies in~\cite{chen2019looks,donnelly2022deformable}, the training procedure of our ST-ProtoPNet consists of 3 stages: 
1) optimisation of the CNN backbone, add-on layers, and prototype layer, using a fixed FC layer initialised with +1.0 and -0.5 for correct and incorrect connection weights, respectively. 
A warm-up of 5 epochs is involved in this stage by updating only the parameters of add-on layers and prototype layer, with a frozen pre-trained CNN backbone. 
2) prototype projection by updating each prototype with its nearest latent training image patch; and
3) optimisation of the FC layer, with an additional $L_1$ regularisation on the incorrect connection weights (initially fixed at -0.5). 
In each stage, we alternate the optimisation of each branch of the ST-ProtoPNet between mini-batches.
Notice that our method only brings marginal extra model parameters, computational complexity, and training time since the CNN backbone is shared and optimised by both branches.

\textbf{Testing.} To exploit the complementary results from both branches of ST-ProtoPNet, its final classification is obtained from the summed logits predicted by the two branches. 
It is worth noticing that this ensemble strategy introduces no loss of interpretablity but improved accuracy.

\vspace{-4pt}
\section{Experiments}
\vspace{-4pt}
\label{sec:experiment}

We perform experiments on three fine-grained classification benchmark datasets: CUB-200-2011 \cite{wah2011caltech}, Stanford Cars \cite{krause20133d}, and Stanford Dogs \cite{khosla2011novel}. 
To achieve fair comparison, we follow previous studies \cite{chen2019looks,wang2021interpretable} by applying offline data augmentations (e.g., random rotation, skew, shear, and left-right flip) on the cropped CUB and cropped Cars datasets (using the bounding boxes provided). 
We also validate our method on the full (i.e., uncropped) CUB and Dogs datasets, and employ the same online data augmentation methods (e.g., random affine transformation and left-right flip) as used in Deformable ProtoPNet \cite{donnelly2022deformable}. 
All images are resized to $224 \times 224$ pixels as network input.

\vspace{-3pt}
\subsection{Experimental Settings}
\vspace{-3pt}
The proposed ST-ProtoPNet method is evaluated on the following CNN architectures: VGG-16, VGG-19, ResNet-34, ResNet-50, ResNet-152, DenseNet-121, and DenseNet-161. 
All CNN backbones are pre-trained on ImageNet~\cite{deng2009imagenet}, except for ResNet-50, which is pre-trained on iNaturalist~\cite{van2018inaturalist} for the experiment on full CUB~\cite{donnelly2022deformable}. 
The add-on layers include two $1 \times 1$ convolutional layers. 
For simplicity, we utilise the same prototype dimension $D=64$ for all CNN backbones on the three datasets. 
For cropped CUB and Cars datasets, following~\cite{wang2021interpretable}, we use 10 prototypes (5 support and 5 trivial) per class and the projection metric in the similarity function ${\rm{sim}}(\cdot,\cdot)$.
In full CUB and Dogs datasets, to ensure comparison fairness with Deformable ProtoPNet~\cite{donnelly2022deformable} that uses 10 $2 \times 2$ (full CUB) and 10 $3 \times 3$ (full Dogs) deformable prototypes per class, we utilise the same total number of prototypes, i.e., 40 $1 \times 1$ (20 support and 20 trivial) for full CUB and 90 $1 \times 1$ (45 support and 45 trivial) for full Dogs.
Also, we employ the cosine similarity in ${\rm{sim}}(\cdot,\cdot)$ and obtain $14 \times 14$ ($\bar{H}=\bar{W}=14$) feature maps by upsampling the original $7 \times 7$ feature maps via a bi-linear interpolation step, as in \cite{donnelly2022deformable}. 
Following previous prototype-based methods \cite{chen2019looks,wang2021interpretable,donnelly2022deformable}, we set $\lambda_1 = 0.8$, $\lambda_2 = 0.48$ and $0.08$ for the support and trivial ProtoPNet branches respectively, $\lambda_4 = 0.001$. We choose $\lambda_3 = 1.0$ with an ablation provided in the supplementary material.

\vspace{-3pt}
\subsection{Interpretability Evaluation} 
\vspace{-3pt}

Rather than performing user-based evaluations~\cite{zunino2021explainable,rymarczyk2022interpretable}, whose results are often subjective and difficult to reproduce \cite{wang2023generalized}, we leverage the annotated object masks to measure interpretability based on the following metrics:

\noindent
\textbf{Content Heatmap (CH)} \cite{pillai2021explainable}: quantifies the percentage of an activation heatmap that lies within the annotated mask. Hence, we expect this metric to be high.

% \noindent
% \textbf{Context Heatmap (CXH)}: This metric quantifies the percentage of a activation heatmap that lies outside the annotated mask. Therefore, we hope this metric to be low.

\noindent
\textbf{Outside-Inside Relevance Ratio (OIRR)} \cite{lapuschkin2016analyzing}: calculates the ratio of mean activation outside the object to mean activation inside the object. 
A low OIRR indicates a method relies more on the object region and less on the context to support its decision, we thus anticipate OIRR to be low. 

\noindent
\textbf{Intersection over Union (IoU)} \cite{chen2018encoder}: measures the mean IoU score, where a threshold of 0.5 is applied on the min-max normalised heatmap to select foreground objects. 

\noindent
\textbf{Deletion AUC (DAUC)} \cite{petsiuk2018rise}: estimates a decrease in the probability of the predicted class as more and more important/activated pixels are removed. 
The area under the probability curve is defined as DAUC. Consequently, a sharp drop and a low DAUC mean better interpretation.

% \noindent
% \textbf{Average Precision (AP)}: As in objection detection task, we also use AP to assess the object localisation performance. 

\noindent
\textbf{Average Inter-class Prototype Distance (AIPD)}, \textbf{Average Inter-class Feature Distance (AIFD)}: we had earlier stated that support prototypes of different classes should be close to each other and trivial prototypes should be far from each other. Thus, we compute the average inter-class cosine distance for prototypes and their nearest local feature representations, respectively. We expect AIPD $<$ AIFD for support ProtoPNet and AIPD $>$ AIFD for trivial ProtoPNet. 

% as close as possible 
% as far as possible 

\vspace{-3pt}
\subsection{Classification Performance}
\vspace{-3pt}

\begin{table*}[]
\setlength{\tabcolsep}{1.0 mm}
\resizebox{\linewidth}{!}{

% \hspace{1.5pt}
\begin{tabular}{c|cccccc|cccccc} 
\hline
\multirow{2}{*}{Method} & \multicolumn{6}{c|}{CUB}                                                                                                          & \multicolumn{6}{c}{Cars}                                                                                                           \\ 
\cline{2-13}
                        & VGG16               & VGG19               & ResNet34            & ResNet152           & Dense121            & Dense161            & VGG16               & VGG19               & ResNet34            & ResNet152           & Dense121            & Dense161             \\ 
\hline
Baseline                & 73.3 ± 0.2          & 74.7 ± 0.4          & 82.2 ± 0.3          & 80.8 ± 0.4          & 81.8 ± 0.1          & 82.1 ± 0.2          & 87.3 ± 0.4          & 88.5 ± 0.3          & \textbf{92.6 ± 0.3} & \textbf{92.8 ± 0.4} & 92.0 ± 0.3          & 92.5 ± 0.3           \\
ProtoPNet \cite{chen2019looks}              & 77.2 ± 0.2          & 77.6 ± 0.2          & 78.6 ± 0.1          & 79.2 ± 0.3          & 79.0 ± 0.2          & 80.8 ± 0.3          & 88.3 ± 0.2          & 89.4 ± 0.2          & 88.8 ± 0.1          & 88.5 ± 0.3          & 87.7 ± 0.1          & 89.5 ± 0.2           \\
TesNet \cite{wang2021interpretable}                 & 81.3 ± 0.2          & 81.4 ± 0.1          & 82.8 ± 0.1          & 82.7 ± 0.2          & 84.8 ± 0.2          & 84.6 ± 0.3          & 90.3 ± 0.2          & 90.6 ± 0.2          & 90.9 ± 0.2          & 92.0 ± 0.2          & 91.9 ± 0.3          & 92.6 ± 0.3           \\
Trivial ProtoPNet       & 80.8 ± 0.2          & 81.2 ± 0.2          & 82.5 ± 0.2          & 83.1 ± 0.3          & 83.9 ± 0.3          & 84.6 ± 0.3          & 90.1 ± 0.2          & 90.7 ± 0.2          & 91.1 ± 0.2          & 91.5 ± 0.2          & 91.4 ± 0.3          & 92.4 ± 0.3           \\
Support ProtoPNet       & 81.7 ± 0.2          & 81.8 ± 0.3          & 83.0 ± 0.1          & 83.6 ± 0.2          & 84.7 ± 0.2          & 85.2 ± 0.3          & 90.9 ± 0.2          & 90.8 ± 0.2          & 91.0 ± 0.2          & 91.8 ± 0.2          & 91.7 ± 0.2          & 92.7 ± 0.3           \\
ST-ProtoPNet (ours)   & \textbf{82.9 ± 0.2} & \textbf{83.2 ± 0.2} & \textbf{83.5 ± 0.1} & \textbf{84.1 ± 0.2} & \textbf{85.4 ± 0.2} & \textbf{86.1 ± 0.2} & \textbf{91.1 ± 0.2} & \textbf{91.7 ± 0.2} & 91.4 ± 0.1          & 92.0 ± 0.2          & \textbf{92.3 ± 0.3} & \textbf{92.7 ± 0.2}  \\
\hline
\end{tabular}

}
\vspace{0.0pt}
\caption{Classification accuracy (\%) on cropped CUB-200-2011 and Stanford Cars by competing methods using different CNN backbones.}
\label{tab:croppedCUBCars}
\end{table*}

Table \ref{tab:croppedCUBCars} presents the classification accuracy (across 5 runs) of our proposed ST-ProtoPNet on cropped CUB and cropped Cars, where the Baseline is represented by non-interpretable black-box CNN models. 
As can be seen, our ST-ProtoPNet outperforms other competing methods across all backbones for the task of bird species classification. 
Also, our method achieves the best results for the car model identification task when using VGG and DenseNet architectures as the CNN backbone. 
In particular, our VGG19-based ST-ProtoPNet reaches an average accuracy of 83.2\% and 91.7\% on CUB and Cars, respectively, surpassing other methods with the most improvements across all backbones. 
% \gustavo{not sure why we focus only on the VGG results here. Is it because it's the most significant we have?  If so, replace more specifically by In particular}. \chong{VGG-19 has the most improvements.}
Moreover, the support ProtoPNet generally performs better than methods utilising only trivial prototypes (e.g., ProtoPNet, TesNet, and Trivial ProtoPNet), showing the importance of learning support prototypes for the interpretable classification. 
It is worth noting that our ST-ProtoPNet produces superior performance over the support ProtoPNet method, indicating that both support and trivial prototypes are useful and can provide complementary information for achieving accurate and interpretable classification. 

% The training of our VGG19-based ST-ProtoPNet takes about 4.5 hours on CUB, without showing obvious training time difference compared to the original ProtoPNet method.
% ST-ProtoPNet increases only a few additional parameters in the two independent add-on layers and fully-connected layers.  

\begin{table*}[]
\setlength{\tabcolsep}{0.6 mm}
\resizebox{\linewidth}{!}{

\begin{tabular}{c|cccccccc|cccccccc} 
\hline
\multirow{2}{*}{Method} & \multirow{2}{*}{\# Prototype} & \multicolumn{7}{c|}{CUB}                                                                                      & \multirow{2}{*}{\# Prototype} & \multicolumn{7}{c}{Dogs}                                                                                       \\ 
\cline{3-9}\cline{11-17}
                        &                            & VGG16         & VGG19         & ResNet34      & ResNet50      & ResNet152     & Dense121      & Dense161      &                            & VGG16         & VGG19         & ResNet34      & ResNet50      & ResNet152     & Dense121      & Dense161       \\ 
\hline
Baseline                & –                          & 70.9          & 71.3          & 76.0          & 78.7          & 79.2          & 78.2          & 80.0          & –                          & 75.6          & 77.3          & 81.1          & 83.1          & 85.2          & 81.9          & 84.1           \\
ProtoPNet \cite{chen2019looks}              & 1×1p, 10pc                 & 70.3          & 72.6          & 72.4          & 81.1          & 74.3          & 74.0          & 75.4          & 1×1p, 10pc                 & 70.7          & 73.6          & 73.4          & 76.4          & 76.2          & 72.0          & 77.3           \\
ProtoPNet \cite{chen2019looks}              & 1×1p, 40pc                 & 72.9          & 74.2          & 74.1          & 84.8          & 76.0          & 76.6          & 78.5          & 1×1p, 90pc                 & 73.9          & 75.3          & 76.1          & 78.1          & 79.7          & 75.4          & 78.8           \\
TesNet \cite{wang2021interpretable}                 & 1×1p, 10pc                 & 75.8          & 77.5          & 76.2          & 86.5          & 79.0          & 80.2          & 79.6          & 1×1p, 10pc                 & 74.3          & 77.1          & 80.1          & 82.4          & 83.8          & 80.3          & 83.8           \\
TesNet \cite{wang2021interpretable}                 & 1×1p, 40pc                 & 77.6          & 79.2          & 76.5          & 87.3          & 80.1          & 80.9          & 81.3          & 1×1p, 90pc                 & 78.5          & 79.6          & 81.2          & 83.3          & 84.5          & 82.1          & 85.2           \\
Deformable ProtoPNet \cite{donnelly2022deformable}  & 2×2p, 10pc              & 75.7          & 76.0          & 76.8          & 86.4          & 79.6          & 79.0          & 81.2          & 3×3p, 10pc                 & 75.8          & 77.9          & 80.6          & 82.2          & 86.5          & 80.7          & 83.7           \\
Trivial ProtoPNet       & 1×1p, 40pc                 & 80.0          & 79.5          & 77.5          & 87.2          & 80.8          & 81.1          & 82.1          & 1×1p, 90pc                 & 78.6          & 80.4          & 82.6          & 85.0          & 87.0          & 82.3          & 85.9           \\
Support ProtoPNet       & 1×1p, 40pc                 & 80.4          & 80.0          & \textbf{78.4} & 87.5          & 80.2          & 81.5          & 82.4          & 1×1p, 90pc                 & 79.0          & 80.6          & 83.0          & 85.1          & \textbf{87.3} & 82.6          & 86.2           \\
ST-ProtoPNet (ours)   & 1×1p, 10pc                 & 76.8          & 77.6          & 77.4          & 86.6          & 78.7          & 78.6          & 80.6          & 1×1p, 10pc                 & 74.2          & 77.2          & 80.8          & 84.0          & 85.3          & 79.4          & 84.4           \\
ST-ProtoPNet (ours)    & 1×1p, 40pc                 & \textbf{81.0} & \textbf{80.2} & 78.2          & \textbf{88.0} & \textbf{81.2} & \textbf{81.8} & \textbf{82.7} & 1×1p, 90pc                 & \textbf{79.1} & \textbf{80.9} & \textbf{83.4} & \textbf{85.7} & 87.2          & \textbf{82.9} & \textbf{86.6}  \\
\hline
\end{tabular}

}
\vspace{0pt}
\caption{Classification accuracy (\%) on full CUB-200-2011 and Stanford Dogs datasets by competing approaches using different CNN backbones, where $\rho_1$×$\rho_2$p denotes the spatial shape of prototypes and $k$pc represents $k$ prototypes per class.}
\vspace{-6pt}
\label{tab:fullCUBDogs}
\end{table*}

Table \ref{tab:fullCUBDogs} shows the classification results on full CUB and full Dogs. 
% We use the same number of prototypes as Deformable ProtoPNet~\cite{donnelly2022deformable}, i.e., 40 $1\times1$ and 90 $1\times1$ prototypes for CUB and Dogs, respectively, for fair comparison. 
In both datasets, the classification accuracy of the original ProtoPNet method is generally worse than the non-interpretable counterpart (Baseline) for many CNN backbones. 
On the other hand, the accuracy by the trivial ProtoPNet and support ProtoPNet are substantially better than those by Baseline, ProtoPNet, and Deformable ProtoPNet. 
However, our ST-ProtoPNet achieves more significant performance gains and exhibits the best accuracy across most backbones, particularly when using a large number of prototypes (i.e., 40 $1\times1$ prototypes per class for CUB and 90 $1\times1$ prototypes per class for Dogs), demonstrating the effectiveness of utilising both the trivial and support prototypes for the interpretable image classification. 
Additionally, when using a smaller number of prototypes, i.e, 10 $1\times1$ prototypes per class, our ST-ProtoPNet still has competitive classification accuracy across multiple backbones.

We further compare our ST-ProtoPNet with other deep-learning methods that can provide different levels of interpretability on CUB, with results shown in Table \ref{tab:interprlevel}, where * and ** denote ensemble of models with different backbones.
% \gustavo{Can we just say "where * and ** denote ensembled models trained with a different number of models"?} \chong{we'd better mention they are from different CNN backbones. }
As evident, an ensemble of three ST-ProtoPNets can achieve high accuracy (87.9\% on cropped images, 88.2\% on full images), outperforming competing methods that are also based on an ensemble of three models (e.g., ProtoTree, TesNet, and ProtoPool).
Moreover, the ensemble of five ST-ProtoPNets outperforms all other competing methods and obtains the best classification accuracy of 88.1\% and 88.4\% on cropped and full CUB images, respectively. 
More results on Cars and Dogs are given in the supplementary material.

\begin{table}[]
\setlength{\tabcolsep}{0.6 mm}
\resizebox{\linewidth}{!}{

\begin{tabular}{llcc} 
\hline
Interpretability level                              & Method                 & \multicolumn{2}{c}{Accuracy (\%)}                    \\ 
\hline
None                                                & B-CNN \cite{lin2015bilinear}                  & 85.1 (b)          & 84.1 (f)                    \\ 
\hline
\multirow{2}{*}{Object-level attention}             & CAM \cite{zhou2016learning}                   & 70.5 (b)          & 63.0 (f)                    \\
                                                    & CSG \cite{liang2020training}                   & \textbf{82.6} (b) & \textbf{\textbf{78.5}}~(f)  \\ 
\hline
\multirow{5}{*}{Part-level attention}               & PA-CNN \cite{krause2015fine}                 & 82.8 (b)          & --                          \\
                                                    & MG-CNN \cite{wang2015multiple}                & \textbf{83.0} (b)          & 81.7 (f)                    \\
                                                    & MA-CNN \cite{zheng2017learning}                & --                & 86.5 (f)                    \\
                                                    & RA-CNN \cite{fu2017look}                & --                & 85.3 (f)                    \\
                                                    & TASN \cite{zheng2019looking}                  & --                & \textbf{\textbf{87.0}}~(f)  \\ 
\hline
\multirow{10}{*}{Part-level attention + Prototypes} & Region \cite{huang2020interpretable}                & 81.5 (b)          & 80.2 (f)                    \\
                                                    & ProtoPNet* \cite{chen2019looks}            & 84.8 (b)          & 81.1 (f)                    \\
                                                    & ProtoTree* \cite{nauta2021neural}            & --                & 86.6 (f)                    \\
                                                    & TesNet* \cite{wang2021interpretable}               & 86.2 (b)          & 83.5 (f)                    \\
                                                    & ProtoPool* \cite{rymarczyk2022interpretable}            & 87.5 (b)          & --                          \\
                                                    & ST-ProtoPNet* (ours)         & 87.9 (b)          & 88.2 (f)                     \\
                                                    & ProtoTree** \cite{nauta2021neural}           & ~--               & 87.2 (f)                    \\
                                                    & Deformable ProtoPNet** \cite{donnelly2022deformable} & ~--               & 87.8 (f)                    \\
                                                    & ProtoPool** \cite{rymarczyk2022interpretable}           & 87.6 (b)          & --                          \\
                                                    & ST-ProtoPNet** (ours) & \textbf{88.1} (b) & \textbf{88.4} (f)                         \\
\hline
\end{tabular}

}
\vspace{0pt}
\caption{Classification accuracy and interpretability level of different methods on CUB-200-2011. “b” and “f” denote the model is trained and tested on cropped and full images, respectively. *: Ensemble of three models. **: Ensemble of five models. }
\label{tab:interprlevel}
\end{table}

\vspace{-3pt}
\subsection{Interpretability Comparison}
\vspace{-3pt}

We assess the model interpretabilty on full CUB using the annotated bird segmentation mask\footnote{http://www.vision.caltech.edu/datasets/}. 
Quantitative results on the test set are given in Table \ref{tab:interpretability}, where all methods are based on the VGG19 backbone. We use GradCAM \cite{selvaraju2017grad} for the non-interpretable baseline. 
For prototype-based methods, we average the activation map of all prototypes of a class to compute the metrics. 
We can see our proposed support ProtoPNet can effectively improve the interpretaiblity in all measures, showing the interpretations produced by our support prototypes are more likely to be object-dependent and focus less on context cues. 
Also, our ST-ProtoPNet method shows better interpretaiblity results than the support ProtoPNet in terms of CH, OIRR, and DAUC. 
Compared with the original ProtoPNet \cite{chen2019looks}, our ST-ProtoPNet obtains significant interpretability improvements.
We show some example activation maps in the supplementary material, and an experiment with prototype pruning is also provided.
% and ablation on independent add-on layers

\begin{table}[]
\setlength{\tabcolsep}{0.4 mm}
\resizebox{\linewidth}{!}{

\begin{tabular}{cccccccc} 
\hline
Metric                  & GradCAM \cite{selvaraju2017grad} & ProtoPNet \cite{chen2019looks} & TesNet \cite{wang2021interpretable} & DefProto \cite{donnelly2022deformable} & TrvProto & SptProto & ST-Proto  \\ 
\hline
CH (\%, $\uparrow$)     & 52.46   & 48.66     & 59.38   & 52.09   & 63.05    & 63.87   & \textbf{66.43}   \\
IoU (\%, $\uparrow$)    & 39.91   & 38.03     & 36.92   & 40.77   & 37.74    & \textbf{42.04}   & 41.05   \\
OIRR (\%, $\downarrow$) & 37.01   & 37.26     & 38.97   & 28.68   & 34.48    & 28.69   & \textbf{28.09}   \\
DAUC (\%, $\downarrow$)  & 7.01    & 7.39      & 5.86    & 5.99    & 6.06    & 5.80    & \textbf{5.74}   \\
\hline
\end{tabular}

}
\vspace{0pt}
\caption{Quantitative interpretability results on full CUB test set. DefProto = Deformable ProtoPNet, TrvProto = Trivial ProtoPNet, SptProto = Support ProtoPNet, ST-Proto = ST-ProtoPNet. 
% \chong{For our ST-ProtoPNet, I use $0.8 \times supportbranch + 0.2 \times trivialbranch$, because we already know the support prototypes is better than the trivial prototypes in terms of interpretaibility (the third last and second last column), what is your opinion for the weighted sum? We can even provide an ablation in the supplement to vary to combination weight.} \gustavo{It's better to add this information to the text, where you say that you average the activation maps of all prototypes, and say that in the supplementary material you show experiments on how you selected the weights for the support and trivial maps.} \chong{discuss later.} 
}
\vspace{-8pt}
\label{tab:interpretability}
\end{table}

Table \ref{tab:proto_feat_dist} presents the computed AIPD and AIFD for the support and trivial ProtoPNet on cropped CUB, using VGG19 and ResNet34 as CNN backbones. As evident, the AIPD is indeed smaller than AIFD for the support ProtoPNet while AIPD is larger than AIFD for the trivial ProtoPNet. This result indicates that the support prototypes of different classes lie closer than their local feature representations and are more inclined to focus on visually similar (i.e., hard-to-learn) object parts of different classes.

\begin{table}[]
\setlength{\tabcolsep}{2. mm}
\resizebox{\linewidth}{!}{

\begin{tabular}{cc|cc|cc|cc} 
\hline
\multicolumn{4}{c|}{VGG19}                                                      & \multicolumn{4}{c}{ResNet34}                                                    \\ 
\hline
\multicolumn{2}{c|}{Support ProtoPNet} & \multicolumn{2}{c|}{Trivial ProtoPNet} & \multicolumn{2}{c|}{Support ProtoPNet} & \multicolumn{2}{c}{Trivial ProtoPNet}  \\
AIPD & AIFD                            & AIPD & AIFD                            & AIPD & AIFD                            & AIPD & AIFD                            \\ 
\hline
 0.7259    & 0.9264                                & 0.9987     & 0.9232                               & 0.6541    & 0.8573                       & 1.000     & 0.9481                                  \\
\hline
\end{tabular}

}
\vspace{0pt}
\caption{Average inter-class prototype distance (AIPD) and average inter-class feature distance (AIFD) for the support and trivial ProtoPNet trained on cropped CUB.}
\vspace{-8pt}
\label{tab:proto_feat_dist}
\end{table}

\vspace{-5pt}
\subsection{Visualisation Analysis}
\vspace{-5pt}

\begin{figure}[t!]
    \centering
    \includegraphics[width=1.00\linewidth]{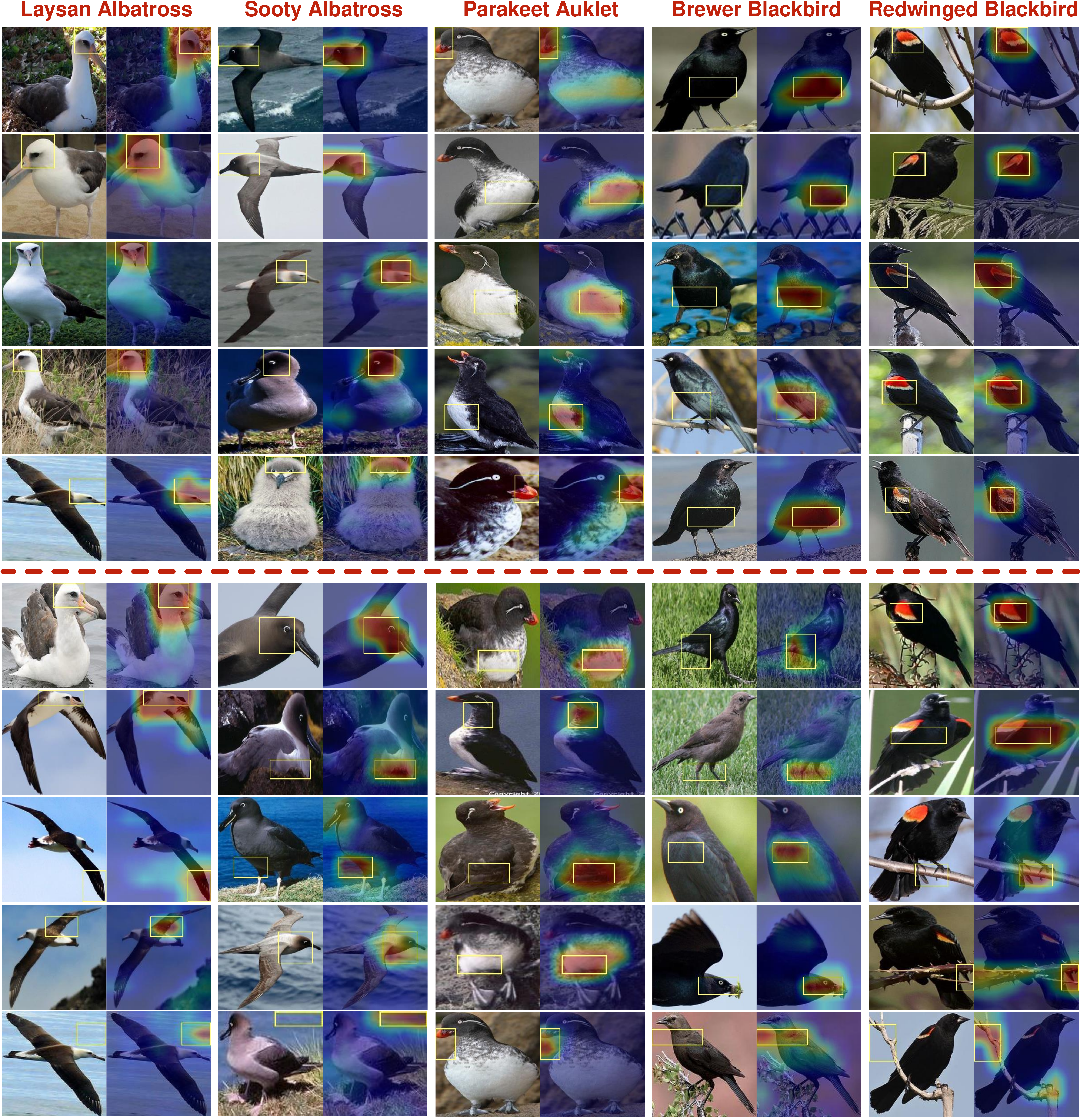}
    \vspace{-12pt}
    \caption{The support (top) and trivial (bottom) prototypes from cropped CUB. In each pair, the first column shows the original image with a prototype marked in a yellow bounding box, the second column is the prototype’s corresponding activation map. }
    \label{fig:visualprototypes}
    \vspace{-16pt}
\end{figure}

\begin{figure}[t!]
    \centering
    \includegraphics[width=1.00\linewidth]{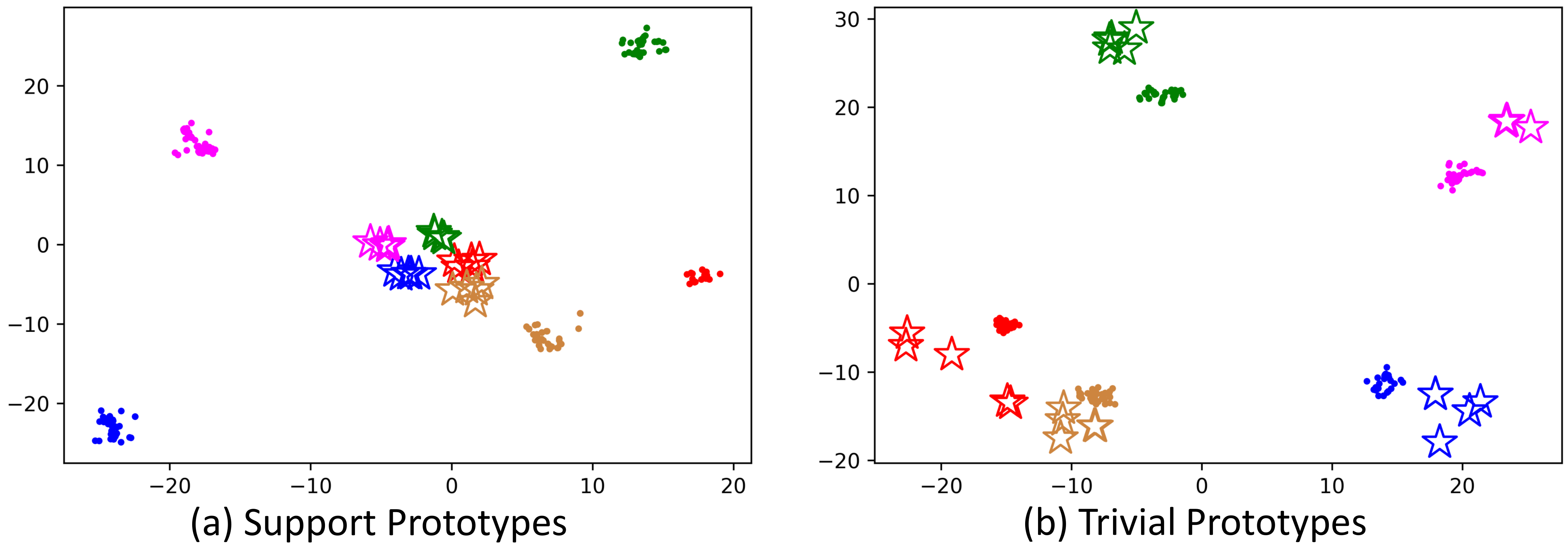}
    \vspace{-13pt}
    \caption{T-SNE results of support and trivial prototypes. The prototypes and their nearest latent training features are marked with stars and dots, respectively. We show results before the prototype projection stage to better visualise the relation between the prototypes and features. Each colour represents a different class. 
    }
    \label{fig:tsne}
    \vspace{-10pt}
\end{figure}

To explore the differences between the support and trivial prototypes, we select 5 categories of birds with visually similar features from cropped CUB to train the support and trivial ProtoPNet methods, with the learned prototypes shown in Fig.~\ref{fig:visualprototypes}.
We notice the support prototypes can capture subtle and fine visual features of different classes and they only focus on relevant bird parts, e.g., head and belly. 
This is reasonable since our algorithm is designed to produce prototypes that are as close as possible to each other, where the image prototypical parts should not only be discriminative but also share visually similar features among classes.
%should share some common features (e.g., head and belly in this case).
By contrast, the trivial prototypes tend to focus not only on the relevant bird parts but also the background regions. 
For example, some trivial prototypes of the Laysan Albatross and Sooty Albatross classes capture the sea surface as they often appear with the sea background. 
We argue that this is because the trivial ProtoPNet may treat the background as an easy pattern to learn, 
%the most easy-to-learn pattern, 
focusing less on the object's visual parts of the class. 
% \gustavo{That argument may be problematic if the reviewer asks: the sea surface is also in the background of Sooty Albatross, so it should be close the the classification boundary between these two classes. The only way out is if we speculate that the sea background seen in both classes look different.}
In Fig.~\ref{fig:tsne}, we show the t-SNE result from the 5-category bird classification, where we note the support prototypes of different classes are located closer to each other, in comparison with the trivial prototypes.

\begin{figure}[t!]
    \centering
    \includegraphics[width=1.0\linewidth]{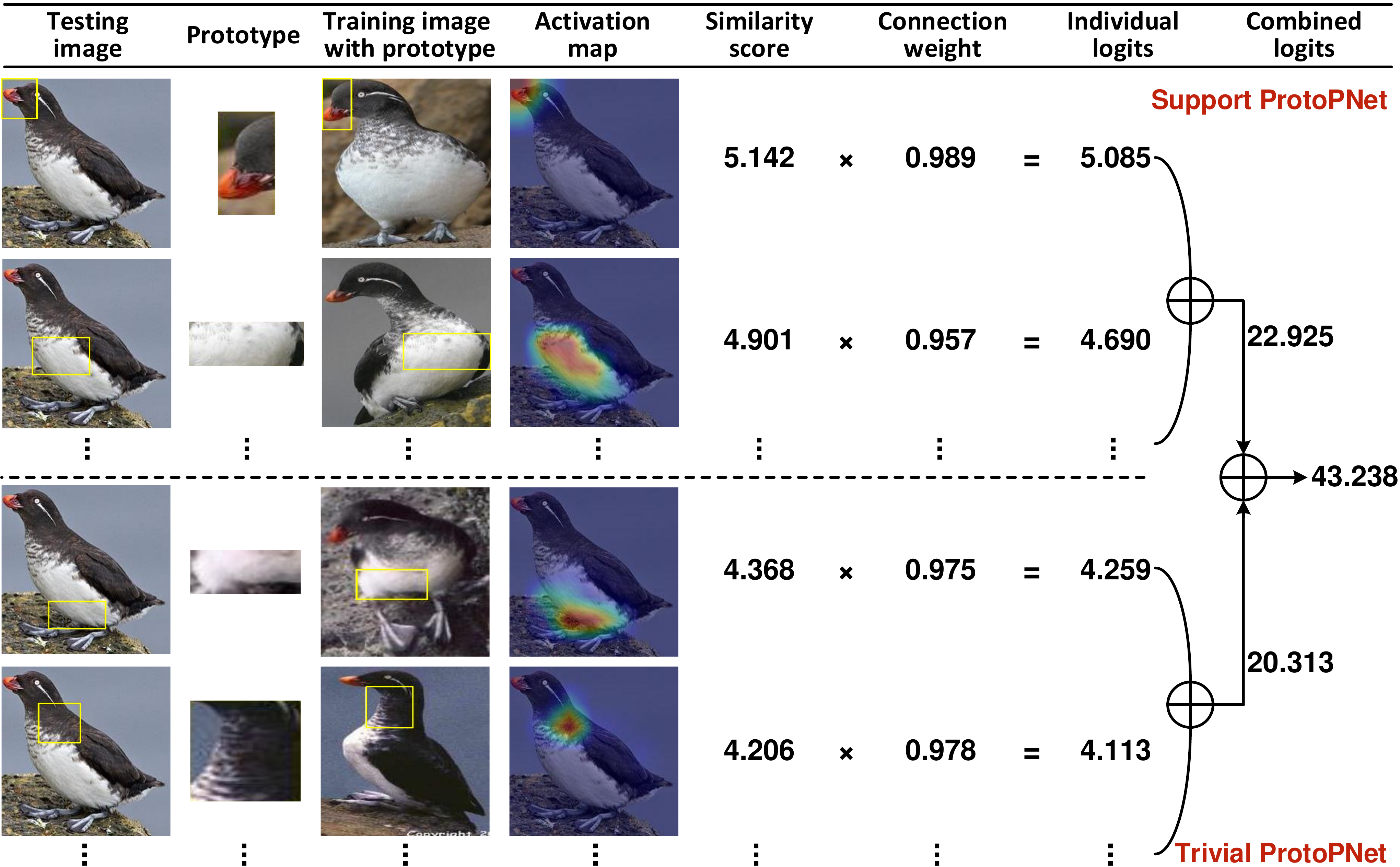}
    % \vspace{-17pt}
    \caption{An example of the interpretable reasoning of our ST-ProtoPNet for classifying a testing Parakeet Auklet image.}
    \label{fig:reasoning}
    \vspace{-1pt}
\end{figure}

Fig.~\ref{fig:reasoning} shows an example of the interpretable reasoning for our ST-ProtoPNet in classifying a testing bird image. 
As evident, each ProtoPNet branch calculates its own classification logits (weighted sum of similarity scores), which is then combined to generate the final prediction. 
Specifically, when classifying a Parakeet Auklet, the support prototypes are quite active on the bird's beak and belly.  
Meanwhile, the trivial prototypes have high activations on the bird’s lower surface and neck. 
In this case, the support ProtoPNet obtains a relatively higher similarity score (22.925), compared with the trivial branch (20.313).
%of the trivial ProtoPNet) between the belly of the bird in the testing image and the belly training prototypes. 
Note that our ST-ProtoPNet exploits both the support and trivial prototypes to capture much richer representations of the object from different perspectives, which enables the realisation of complementary interpretations.
% the final decision in a complementary way. 
% Notably, the ensemble classification improves interpretability and accuracy using richer representations for the target object. 
% The support and trivial ProtoPNets make complementary predictions contributing to the final classification decision. 
% More examples of visual prototypes and interpretable reasoning on Cars and Dogs are provided in the supplementary material. 
More examples on Cars and Dogs are shown in the supplementary material.

\vspace{-3pt}
\subsection{Ablation Study}
\vspace{-3pt}

\textbf{The closeness and discrimination losses.} 
To validate the effectiveness of our proposed closeness loss in Eq.~\eqref{eq:support_cls} and discrimination loss in Eq.~\eqref{eq:ell_dis}, 
we first conduct ablation studies on full CUB and full Dogs by using ResNet50 and ResNet34 as the CNN backbone, respectively. 
Results are listed in Table~\ref{tab:ablationLoss}. 
We can observe that both the closeness and discrimination losses can improve the accuracy, compared with the Baseline ProtoPNet method trained with only clustering, separation, and orthonormality losses. 
Note that the closeness loss introduces a larger performance improvement, which is attributed to the learning of support (i.e., hard-to-learn) prototypes.

% \gustavo{Should we present the whole ablation study with the other losses (CT, SP, ORT)? Maybe we can add this result to the supplmentary material and say that it is not important to add it here since our main interests are in the DSC and CLS losses.} \chong{I think we may not need do it. Because the CT and SP are proposed by the original work ProtoPNet which is necessary to regularise the ProtoPNet's training. The ORT is proposed by TesNet which has conducted an ablation in that paper.}

\begin{table}[]
\setlength{\tabcolsep}{3.4 mm}
\resizebox{\linewidth}{!}{

\begin{tabular}{cccccccc} 
\hline
\multirow{2}{*}{Method} & \multirow{2}{*}{$\ell_{ct}$} & \multirow{2}{*}{$\ell_{sp}$} & \multirow{2}{*}{$\ell_{ort}$} & \multirow{2}{*}{$\ell_{dsc}$} & \multirow{2}{*}{$\ell_{cls}$} & \multicolumn{2}{c}{Accuracy (\%)}  \\
                        &                                                  &                                                  &                                                   &                                                   &                                                   & CUB & Dogs                    \\ 
\hline
Baseline     & $\checkmark$                     & $\checkmark$                    & $\checkmark$                                      &                                                   &                                          & 86.5    & 80.9                       \\
Trivial ProtoPNet       & $\checkmark$                     & $\checkmark$                    & $\checkmark$                                      & $\checkmark$                                      &                                          & 87.2    & 82.6                        \\
Support ProtoPNet       & $\checkmark$                     & $\checkmark$                    & $\checkmark$                                      &                                                   & $\checkmark$                             & 87.5    & 83.0                        \\
ST-ProtoPNet (ours)      & $\checkmark$                     & $\checkmark$                    & $\checkmark$                                      &                       $\checkmark$                                 & $\checkmark$                             & 88.0    & 83.4                        \\
\hline
\end{tabular}

}
\vspace{0pt}
\caption{Ablation analysis of the 
closeness loss $\ell_{cls}$ in Eq.~\eqref{eq:support_cls} to learn support prototypes, and 
discrimination loss $\ell_{dsc}$ in Eq.~\eqref{eq:ell_dis} to learn trivial prototypes on full CUB-200-2011 and Stanford Dogs.
%\gustavo{We have space to add the results from Table 2, but the results don't help us on CUB... I guess this can be fixed if we show results with a different backbone.  In any case, I think we should show the results of ST-ProtoPNet.} \chong{now we use ResNet50 for CUB, ResNet34 for Dogs.}
}
\vspace{-10pt}
\label{tab:ablationLoss}
\end{table}

\textbf{Combining Support and Trivial Prototypes.} 
We also investigate the importance of integrating the two complementary sets of support and trivial prototypes for improved classification. 
To achieve this, we first train a two-branch model where both branches learn the same type of prototypes and the final result is produced by the ensemble of them (Trivial Ensemble and Support Ensemble). 
Besides, for our ST-ProtoPNet, we also provide results of its individual branches (Trivial Branch and Support Branch).
Table~\ref{tab:ablationEnsemble} shows the experimental results on cropped CUB. 
We can notice that combining the two different types of prototypes (ST-ProtoPNet) achieves superior performance over combining only the same type of prototypes (Trivial Ensemble and Support Ensemble), indicating that our performance improvements are from not only the ensemble strategy but also the two complementary sets of prototypes. 
Also, ST-ProtoPNet indeed exhibits higher accuracy than its individual branches, further verifying that the results from the two branches are complementary, and the combination of them is effective to improve the final classification accuracy.

\begin{table}[]
\setlength{\tabcolsep}{0.7 mm}
\resizebox{\linewidth}{!}{

\begin{tabular}{ccccccc} 
\hline
Method              & VGG16               & VGG19               & ResNet34            & ResNet152           & Dense121            & Dense161             \\ 
\hline
Trivial Ensemble    & 81.4 ± 0.3          & 81.8 ± 0.2          & 82.7 ± 0.2          & 83.2 ± 0.3          & 84.4 ± 0.2          & 85.0 ± 0.3           \\
Support Ensemble    & 82.1 ± 0.2          & 82.4 ± 0.3          & 83.0 ± 0.2          & 83.7 ± 0.3          & 84.8 ± 0.2          & 85.5 ± 0.2           \\
Trivial Branch      & 81.0 ± 0.2          & 81.1 ± 0.3          & 82.4 ± 0.2          & 82.9 ± 0.3          & 84.1 ± 0.3          & 84.8 ± 0.3           \\
Support Branch      & 81.5 ± 0.3          & 81.8 ± 0.3          & 82.8 ± 0.2          & 83.4 ± 0.3          & 84.6 ± 0.2          & 85.4 ± 0.2           \\
ST-ProtoPNet (ours) & \textbf{82.9 ± 0.2} & \textbf{83.2 ± 0.2} & \textbf{83.5 ± 0.1} & \textbf{84.1 ± 0.2} & \textbf{85.4 ± 0.2} & \textbf{86.1 ± 0.2}  \\
\hline
\end{tabular}

}
\vspace{0pt}
\caption{Ablation study of the combination of support and trivial prototypes for improved classification on cropped CUB-200-2011.}
\label{tab:ablationEnsemble}
\vspace{-10pt}
\end{table}

\vspace{-6pt}
\section{Conclusion and Future Work}
\vspace{-5pt}

In this paper, we proposed the ST-ProtoPNet to exploit both support (i.e., hard-to-learn) and trivial (i.e., easy-to-learn) prototypes, where the two sets of prototypes can provide complementary results for the interpretable image classification. 
Our ST-ProtoPNet is a general approach that can be easily applied to existing prototype-based interpretable models. 
One limitation for our method is that we empirically adopt the same number of support and trivial prototypes and the same total number of prototypes for each class. 
Considering the different learning difficulties and imbalanced training samples among classes in other real-world datasets, e.g., ImageNet \cite{deng2009imagenet}, a better way to adaptively learn a flexible number of support and trivial prototypes is needed and deserves to be further investigated in our future work. 
% \sout{In this paper, we mimic the behaviour of the support vectors of SVM classifier to obtain the support prototypes by forcing them to be as close as possible to the classification boundary. 
% In the future, we will explore to formulate the learning of prototypes with gradient-based kernel computations, e.g., neural tangent kernel \cite{jacot2018neural} and path kernel \cite{domingos2020every}.}
Moreover, given that we mimic the behaviour of the support vectors of SVM classifier to obtain the support prototypes by forcing them to be as close as possible to the classification boundary, we plan to develop new methods to learn prototypes with gradient-based kernel techniques, e.g., neural tangent kernel~\cite{jacot2018neural} and path kernel~\cite{domingos2020every}.

\vspace{10pt}
\noindent
\textbf{Acknowledgements}.
This work was supported by funding from the Australian Government under the Medical Research Future Fund - Grant MRFAI000090 for the Transforming Breast Cancer Screening with Artificial Intelligence (BRAIx) Project, and the Australian Research Council through grant FT190100525.

%%%%%%%%% REFERENCES

{\small
\bibliographystyle{ieee_fullname}
\bibliography{refs}
}

\end{document}